\def\year{2021}\relax
\documentclass[letterpaper]{article} % DO NOT CHANGE THIS
\usepackage{aaai21}  % DO NOT CHANGE THIS
\usepackage{times}  % DO NOT CHANGE THIS
\usepackage{helvet} % DO NOT CHANGE THIS
\usepackage{courier}  % DO NOT CHANGE THIS
\usepackage[hyphens]{url}  % DO NOT CHANGE THIS
\usepackage{graphicx} % DO NOT CHANGE THIS
\usepackage{natbib}  % DO NOT CHANGE THIS AND DO NOT ADD ANY OPTIONS TO IT
\usepackage{caption} % DO NOT CHANGE THIS AND DO NOT ADD ANY OPTIONS TO IT
\usepackage[switch]{lineno} 

\usepackage[ruled,vlined,linesnumbered]{algorithm2e}
\usepackage{multirow}
\usepackage{multicol}
\usepackage{amssymb}
\usepackage{booktabs}
\usepackage{subcaption}

\nocopyright
\title{Worst-Case-Aware Curriculum Learning for Zero and Few Shot Transfer}

\author{Sheng Zhang\textsuperscript{\rm 1,2}, Xin Zhang\textsuperscript{\rm 1}, Weiming Zhang\textsuperscript{\rm 1}, Anders Søgaard\textsuperscript{\rm 2} \\
}
\affiliations{
    \textsuperscript{\rm 1}National University of Defense Technology  \\
    \textsuperscript{\rm 2}University of Copenhagen\\
    zhangsheng@nudt.edu.cn,
    ijunzhanggm@gmail.com,\\
    wmzhang@nudt.edu.cn,
    soegaard@di.ku.dk
}

\begin{document}
% \linenumbers  %

\maketitle

\begin{abstract}
Multi-task transfer learning based on pre-trained language encoders achieves state-of-the-art performance across a range of tasks. Standard approaches implicitly assume the tasks, for which we have training data, are equally representative of the tasks we are interested in, an assumption which is often hard to justify. This paper presents a more agnostic approach to multi-task transfer learning, %, relying on $\alpha$-ball 
which uses automated curriculum learning to minimize a new family of worst-case-aware losses across tasks. Not only do these losses lead to better performance on outlier tasks; they also lead to better performance in zero-shot and few-shot transfer settings.
\end{abstract}

\section{Introduction}

Multi-task learning is the problem of minimizing the average error across $n$ tasks, as measured on held-out samples, and motivated by the observation that sometimes learning a single model with partially shared parameters performs better than $n$ single-task models. In the learning-to-learn setting, we worry about our error on a task $n+1$. Both of these settings apply to randomly initialized base learners, as well as architectures pre-trained on yet another task(s). In learning-to-learn, the new task $n+1$ is assumed to come from an ambiguity set defined by the $n$ tasks.

Unsurprisingly, most approaches to multi-task learning minimize the average loss across the training samples available for these tasks. This does not always lead to the best solution, however, since the relations between loss and error may differ across tasks. Several off- and online methods for normalizing these relations have been proposed \citep{chen2018gradnorm,bronskill2020tasknorm}, but even with this, minimizing average loss across tasks has two disadvantages: (a) Performance on outlier tasks may be very poor \cite{NIPS2010_4150,pmlr-v37-hernandez-lobatoa15}; and (b) in the learning-to-learn setting, minimizing average loss is only optimal if the task selection is unbiased \citep{DBLP:journals/neco/Wolpert96,oren-etal-2019-distributionally,DBLP:journals/corr/abs-2003-06054}. 

Minimizing the worst-case loss across tasks instead of the average loss, in theory solves these two problems, which is why this approach is popular in algorithmic fairness \citep{pmlr-v80-hashimoto18a} and domain adaptation under covariate shift assumptions \citep{Duchi:ea:20}. In some multi-task settings, it is possible to directly modify the loss that is minimized in multi-task learning \citep{mehta2012minimax}, but this is for example not possible in the common approach to multi-task learning where each batch is sampled from one of $n$ tasks at random \citep{Caruana:97,Baxter:00}. We present a more general approach to multi-task learning with worst-case-aware loss minimization, instead relying on automated curriculum learning \citep{graves2017automated}. 

\paragraph{Contributions} We present an automated curriculum learning approach to robust multi-task transfer learning. Our approach is general and parameterizes a family of worst-case-aware objectives, with minimax and loss-proportional minimization at the two extremes. In a series of experiments on the GLUE multi-task benchmark \citep{wang2018glue}, we show that several of these objectives lead to better performance on the benchmark itself, but more importantly, also lead to much better (zero-shot {\em and} few-shot) generalization to other out-of-domain data sets. %Finally, we show that the shared models learned using worst-case-aware curriculum learning also perform better in learning-to-learn settings. 

\section{Worst-Case-Aware Multi-Task Curriculum Learning}
\begin{figure*}[htbp]
\centering
\includegraphics[width=12cm]{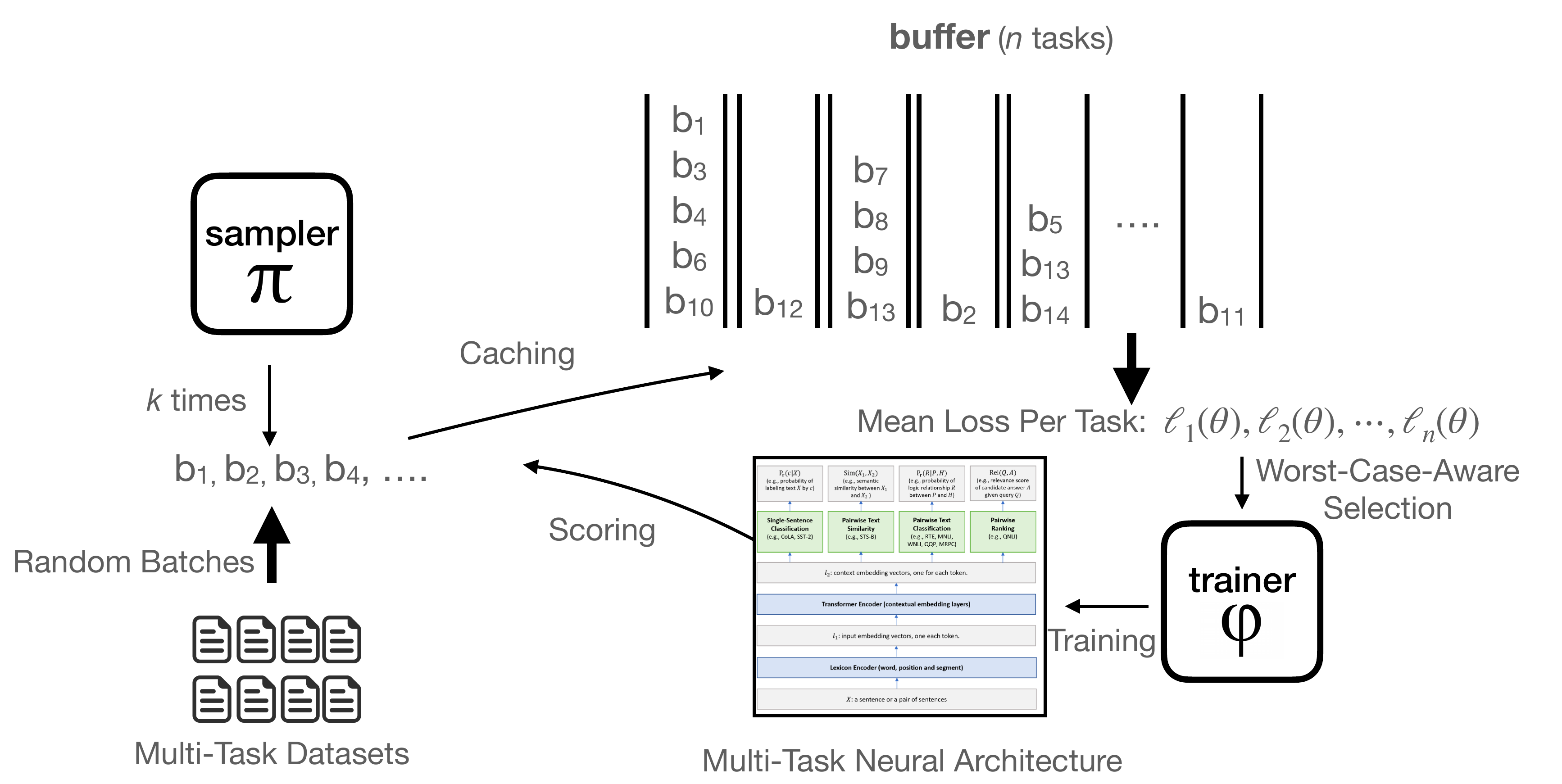}
\caption{{\sc The Architecture of Worst-Case-Aware Multi-Task Curriculum Learning.} The architecture consists of the sampler, buffer, trainer and the multi-task model. The trainer samples data batches for $k$ times at each round, evaluated by the neural architecture and then cached in the buffer. The buffer consists of $n$ first-in first-out queues. The trainer adopts worst-case-aware strategy to select the task according to average loss in the buffer for the training of the neural architecture.}
\label{fig:framework}
\end{figure*}

In multi-task learning, we minimize a loss over a set of $n$ related tasks:

\begin{equation}
    \min_\theta \ell(\theta) = \min_\theta f_{i\leq n}\ell_i(\theta)
\end{equation}
where $\ell_i(\theta)$ is the loss of the $i$-th task, and $f$ is some function combining the task-specific losses. Most multi-task learning algorithms simply use $\sum(\cdot)$ for $f(\cdot)$, but often task losses are weighted to compensate for differences in data set sizes and complexities \citep{Stickland:Murray:19}. \citet{mehta2012minimax}, in contrast, explore the $\left\|\ell_i(\theta)\right\|^j$ family of losses for multi-task learning, i.e., of minimizing the $\ell_i$-norm of the loss vector, with $\ell_1$ equivalent to the standard average loss, and $\ell_\infty$ corresponding to worst-case minimization over the $n$ tasks. 

If the tasks and datasets are sampled {\em i.i.d.}, minimizing $\sum_{i\leq n}\ell_i(\theta)$ or $\left\|\ell_1(\theta)\right\|^j$ makes sense, but if not, minimizing average error is unlikely to generalize well. In practice, moreover, some tasks may be easier to learn than others, and their loss may decrease quickly, whereas the loss for other tasks will decline more gradually. 
Common sampling techniques will likely result in overfitting on some tasks and under-fitting on others. Finally, optimizing for one task may impair performance of another task. Minimizing the average loss over $n$ tasks leads to better average performance (if data is {\em i.i.d}), but provides no guarantees for worst-case performance. %There may be a situation when the model is training, it will sacrifice the performance of the worst task in order to further improve the better task, even though the summed loss value is decreasing.
All of this motivates more dynamic training strategies that pay attention to tasks on which our joint model performs poorly. We refer to such training strategies as {\em worst-case-aware} strategies. 

We argue it makes more sense to minimize a worst-case loss when the task selections or the datasets are biased, and all we can assume is that the tasks at hand are members of an ambiguity set from which future tasks will also be sampled: Assume, for example, that tasks are sampled from an unknown $\alpha$-ball of distributions with uniform probability \citep{DBLP:journals/corr/abs-1908-08729,yue2020linear}. %Since our task sample may be biased, we cannot estimate the probability of tasks.
We can assume that each observed task is within the $\alpha$-ball. In other words, our best estimate of the center of the $\alpha$-ball is not, for example, the Gaussian mean over tasks, but simply their centroid. Worst-case minimization minimizes the expected worst-case loss within this $\alpha$-ball. 

The approach in \citet{mehta2012minimax} is based on computing norms over batches composed of data from multiple tasks. Such an approach would interact heavily with batch size and batch normalization, which are key hyper-parameters in deep learning. We instead propose a much simpler wrapper method, in which the departure from the standard multi-task objective is delegated to a multi-armed bandit used for sampling data batches during training. The multi-armed bandit is trained to optimize a worst-case aware loss. We call this {\em worst-case-aware multi-task curriculum learning}. 

\subsection{Architecture} 

In the remainder of this section, we describe our approach in detail. Our family of loss functions will be different from \citet{mehta2012minimax}. Instead of introducing a continuum from $\ell_1$-loss to $\ell_\infty$-loss, we introduce a continuum from (minimax) $\ell_\infty$-loss to loss-proportional loss, i.e., a stochastic, adaptive form of worst-case optimization. The entire continuum is worst-case-aware and, from an optimization perspective, represents degrees of stochastic exploration.  Our overall architecture is presented in Figure \ref{fig:framework}. 

Our architecture consists of four parts: the sampler, the buffer, the trainer, and the model itself. The {\bf sampler}, i.e., a multi-armed bandit, selects data batches from the $n$ tasks by revising an automated curriculum learning policy, and the losses associated with these batches are then evaluated by the multi-task learning {\bf model}. The {\bf buffer} maintains $n$ fixed-length queues to cache the sampled data and their loss values according to the current model. The {\bf trainer} chooses a task that is worth optimizing for the multi-task learning model to update the parameters of the model and sends rewards to the sampler. We describe our approach to curriculum learning in detail in the next section. Since most state-of-the-art architectures in natural language processing rely on pre-trained language encoders, we will assume the multi-task learning model is, in part, initialized by pre-training with a masked language modeling objective \citep{devlin-etal-2019-bert}. Specifically, we adopt MT-DNN \citep{liu2019multi} as our multi-task learning architecture; see paper for details. We will evaluate our $n$-task fine-tuned model on the $n$ tasks, as well as new tasks ($n+1$), with and without additional data for fine-tuning. We refer to the latter scenarios as {\em few-shot and zero-shot transfer learning}. We will not explore the impact of pre-training on the learned curricula in this paper, but preliminary experiments suggest that effects of worst-case aware learning on generalization are even stronger in randomly initialized networks.  

\begin{algorithm}[htbp]

\SetAlgoLined

 %Calculate $\pi^{(t-1)}$ for each task\;
%  $Q$: a buffer of $n$ queues \;
 \For{$\tau=1,2,\cdots,k$}{
 % May be change t to \tau here
    Take action $a_\tau=i$ by sampling $i\sim \pi^{(t)}$ \;
    \label{alg:sample:action}
    Select data $B_i^{(\tau)}$ and calculate loss $\ell_i^{(\tau)}$ \; 
    \label{alg:sample:select}
    Push tuple $(B_i^{(\tau)},\ell_i^{(\tau)})$ to queue $Q_i$ \; 
    \label{alg:sample:push}
 }
 $\tilde{i} = $ Trainer.ChosenIndex($Q$) \;
 \For{$i=1,2,\cdots,n$}{
 \If{task $i \in [a_1,a_2,\cdots,a_k]$}{
        Calculate $r_i^{(t)}$ according to $\tilde{i}$ \;
        Update $w_i^{(t+1)}$ using Equation (\ref{eqa:w_update}) \; \label{alg:sample:update}
    }
 }
 Empty($Q_{\tilde{i}}$)

\caption{%ACLSampler.
%{\bf sampler}($n$,$w$,$t$,$k$,$Q$)}
{\bf sampler}($\pi$,$w$,$Q$)}
\label{alg:sample}
\end{algorithm}

\subsection{Worst-Case-Aware Curriculum Learning}

The performance of a multi-task neural network trained on data from $n$ tasks on task $i$, is sensitive to {\em how often} and {\em when} the network saw data sampled from task $i$ \citep{Glover:Hokamp:19,Stickland:Murray:19}. The standard approach is to sample batches uniformly at random from the $n$ tasks. %In multi-task learning, the order of the data fed to the model from different tasks influences the learning curve and the performance of multi-task model. The most common approach in practice is to use a random strategy to randomly select data batches according to the size of the dataset. After one epoch, all samples from each tasks are fed into the model once. However, this strategy ignores the different learning difficulties between tasks, and usually results in that the model puts more strength to tasks with larger data volume and ignores smaller ones.
In this paper, we use a multi-armed bandit with $n$ arms \citep{bubeck2008pure} -- one for each task -- to learn curricula in an online fashion. Our approach is loosely inspired by  \citet{graves2017automated}.  %Suppose we have $N$ tasks, each time we sample a data batch of one task from $N$ tasks by analogy with choosing one arm from the $N$-arm bandit.

%The multi-armed bandit is used to learn an adaptive policy. 
Our algorithm is presented in Algorithm~1 and consists of three steps: 

\paragraph{Step 1: Populating the Buffer} This is the step in lines 2--6 of Algorithm~1. The sampler selects a sequence of $k$ arms (out of a total of $n$ arms, and over $R$ rounds) leading to a total of $kR$ steps with  ($1\leq a_\tau \leq n$): $a_{1^1} \ldots a_{1^k} \ldots a_{R^1},\ldots a_{R^k}$. In the $t$th step, if the $i$th arm is selected, a batch of data $B_i^{(\tau)}$ -- or simply $b_\tau$ --  is sampled from task $i$ (line \ref{alg:sample:action}). Selecting the $i$th arm in the $\tau$th action yields a pay-off $r_\tau$ by evaluating the loss $\ell_i^{(\tau)}$ of the multi-task model on the data batch (line \ref{alg:sample:select}). The batch $b_t$ and its loss $\ell_i^{(\tau)}$ are stored in the buffer, i.e., $n$ queues of maximum length $m$. Specifically, $(b_\tau,\ell_i^{(\tau)})$ is pushed to the queue $Q_i$ (line \ref{alg:sample:push}). The buffer thus contains data instances across $n$ queues that are deemed good under $\phi$, from which the multi-task trainer selects its training data by either sampling a task $i$ with likelihood proportional to the average loss in $Q_i$, or simply the task with the highest loss. See Step 2 for details. The arms are selected stochastically according to a policy $\pi$.  %We present our sampling algorithm in Algorithm \ref{alg:sample}.

This procedure is repeated $k$ times at each round, and the sampler updates its policy after receiving rewards from the trainer. 

\paragraph{Step 2: Worst-Case-Aware Training}

%During training, we minimize the loss by updating the model's parameters. % according to the error gradients with back-propagation. 
%For multi-task learning, it is a multi-objective optimization problem, and the model needs to be updated with multiple losses. 
In multi-task learning with $n$ tasks, it is common practice to minimize the sum of losses across the $n$ tasks; i.e., $\min_{\theta}\ell(\theta)=\min_{\theta}\sum_{i\leq n}\ell_i(\theta)$. This is typically implemented by uniformly sampling data across tasks during training. Another approach is to sample size-proportionally \cite{Glover:Hokamp:19,Stickland:Murray:19}, i.e., to rely on a weighted sum, $\sum_{i\leq n}N_i\ell_i(\theta)$, where $N_i$ is the size of the training set for task $i$.  %Another method is to randomly sample one data batch from a task at a time and update parameters of the network according to loss of that task. 
%Therefore, to overcome the weakness mentioned above, we proposed a training strategy for our trainer, which pays more attention to the worst performing task. 
We instead present a general class of worst-case-aware minimization strategies parameterized by the parameter $\phi\in[0,1]$. Our worst-case-aware minimization strategy can be defined as:

\begin{equation}
    \min \ell(\theta) %= \min \mathcal{L}_{\tilde{j}} 
    =\left\{
\begin{array}{rl}
\min \: \max_i (\ell_i(\theta)), &  p < \phi \\
\\
\min \ell_{\tilde{i}}(\theta), &  p \geq \phi, \tilde{i} \sim P_\ell \\
\end{array} \right.
\end{equation}
where $p$ is a randomly generated real number in $[0,1)$; $P_\ell$ is the probability distribution of the current, normalized task losses, i.e., $\frac{\ell_i}{\sum_{j\leq n}{\ell_j}}$; and $\tilde{i}$ is the index of the task that is chosen by our strategies, either because it is the task with the highest loss (when $p<\phi$), or when it is stochastically sampled with the probability of its normalized task loss (when $p \geq \phi$). If $\phi$ equals $1$, our trainer just optimizes for the worst-case task. If $\phi$ equals $0$, the strategy samples loss-proportionally. In practice, the loss of task $i$ is approximated by the average loss of the batches for that task in the buffer.\footnote{The batches in the buffer are not re-evaluated once they are in, in order to speed up training. The loss of new batches is always computed using the current model, and old batches that are not selected are eventually replaced with batches with loss values based on recent models.}% of our algorithm Data batches in the buffer are not re-evaluated considering the time-consuming. Experiments also show no significant difference for performance with re-evaluation.}.  

%In our algorithm, the loss function of task $j$ can be defined as the average loss of all data batches in its corresponding buffer.

\begin{equation}
    \ell_i(\theta) = v_i \cdot \frac{1}{|Q_i|}\sum_{\ell_i^{(j)}(\theta) \in Q_i}(\ell_i^{(j)}(\theta))
\end{equation}
where $v_i$ is the hyper-parameter representing the weight of the task, which can be simply all set to 1 if tasks are treated equally.

%The buffer $Q$ is composed of $N$ first-in first-out queues. The maximum length of each queue is $m$. 
%Inspired by Graves et al. \cite{graves2017automated}, we use the improved EXP3 algorithm to update the sampling strategy. Unlike the vanilla exp3 algorithm, which updates after each action, 
%We update $\pi$ after $k$ actions:[

\begin{table*}[htbp]
\centering
\resizebox{\textwidth}{!}{%
\begin{tabular}{lccccccc}
\toprule
\textbf{Dataset} & \textbf{\#Train}  & \textbf{\#Test} & \textbf{\#Class}    & \textbf{Metrics}  & \textbf{Task} &\textbf{Domain} & \textbf{Classifier}\\
\midrule
\multicolumn{8}{c}{\textbf{GLUE tasks}}  \\ \midrule
CoLA     & 8.5k    & 1k     & 2   & Matthews corr. & Acceptability & General & -\\
SST-2    & 67k     & 1.8k   & 2  & Accuracy  &  Sentiment  &  Movie Reviews&-\\
MRPC     & 3.7k    & 1.7k   & 2   & Accuracy/F1 &  Paraphrase  & News & - \\
STS-B    & 7k      & 1.4k   & 1 & Pearson/Spearman corr. & Sentence Similarity & General & - \\
QQP      & 364k    & 391k   & 2  & Accuracy/F1  & Paraphrase  & Community QA & - \\
MNLI     & 393k    & 20k    & 3   & Accuracy  & NLI & General & -\\
QNLI     & 105k    & 5.7k   & 2  & Accuracy  & NLI/QA & Wikipedia & - \\
RTE      & 2.5k    & 3k     & 2   & Accuracy  &  NLI & Wikipedia & - \\
% WNLI     & 634     & 146    & 2 & Accuracy  &  NLI  & Fiction Books & - \\

\midrule

\multicolumn{8}{c}{\textbf{Transfer learning tasks}}  \\ \midrule
IMDB & 25k &  25k  &  2   &  Accuracy  &  Sentiment  & Movie Reviews  & SST-2 \\
% MR &  -  &    10.6k    & 2 &  Accuracy   &  Sentiment & Movie Reviews   & SST-2 \\
% CR &  -    &   3.8k    &  2 &  Accuracy  &  Sentiment & Customer Reviews   & SST-2\\
Yelp-2 & 56k  & 38k   & 2   & Accuracy &   Sentiment  & Yelp Reviews & SST-2 \\
SICK-R &  5k   & 5k  &  1  & Pearson/Spearman corr.  & Sentence Similarity  & General  & STS-B \\
SICK-E  &  5k   & 5k   & 3  &  Accuracy   & NLI & General & MNLI \\
SNLI     & 55k & 10k   &  3  &   Accuracy     &  NLI   & General & MNLI \\
SciTail     & 24k &  2.1k &    2   &  Accuracy    &  NLI  & Science  & RTE \\
WikiQA     &20k  & 6k &     2  &   MAP/MRR  & QA  & Wikipedia &  QNLI \\
\bottomrule  \\
\end{tabular}
}
\caption{{\sc Dataset Characteristics.} STS-B and SICK-R are regression tasks, while the rest are classification tasks. %As MR and CR datasets do not have official train/test split, we use all samples for evaluation. 
We use SciTail's development set for evaluation, since the test set labels are not public available. IMDB: \cite{maas2011learning};% MR: \cite{Pang+Lee:05a}; CR: \cite{hu2004mining};
Yelp-2: \url{https://www.kaggle.com/yelp-dataset/yelp-dataset}; SICK-R/E: \cite{marelli2014sick}; SNLI: \cite{bowman-etal-2015-large}; SciTail: \cite{khot2018scitail}; WikiQA: \cite{yang-etal-2015-wikiqa}.}
\label{tab:datasets}

\end{table*}

\paragraph{Step 3: Updating $\pi$} Since rewards will change dynamically throughout learning, we adopt the fixed share method \citep{journals/ml/HerbsterW98} to mix in weights additively. The sampling policy $\pi$ in $t$-th round is: % can be defined as:%]%. $\pi$ is defined by the weight $w$, and $\epsilon$-greedy is introduced,

\begin{equation}
    \pi_i^{(t)} = (1-\gamma)\cdot\frac{w_i^{(t)}}{\sum_{j\leq n}{w_j^{(t)}}} + \frac{\gamma}{n}
\end{equation}
where $\gamma$ is an egalitarianism factor which balances the explore-exploit trade-off.

If the arm $i$ is selected in a round with $k$ actions, 

\begin{equation}
r_i^{(t)} = \left\{
\begin{array}{rcl}
\frac{|Q_i^{(t)}|-|Q_i^{(t-1)}|}{\max_j(|Q_j^{(t)}|-|Q_j^{(t-1)}|)}, & & i=\tilde{i} \\ 
\\
-\frac{|Q_i^{(t)}|-|Q_i^{(t-1)}|}{\max_j(|Q_j^{(t)}|-|Q_j^{(t-1)}|)}, & & i \neq \tilde{i}\\
\end{array} \right.
\end{equation}
where $|\cdot|$ means the count of the queue, and $\tilde{i}$ is the index of the task that is chosen by the trainer. Note the reward $r_i^{(t)}$ does not depend directly on the  loss $\ell_i$, but the losses $\ell$ are used to select $\tilde{i}$ in Step 2, which subsequently determines the reward. 

%Assuming that the feedback obtained by the selected $j$-th arm at time step $t$ is
With $r_i^{(t)}$, $w_i^{(t+1)}$ is then updated (Algorithm \ref{alg:sample}, line \ref{alg:sample:update}):

\begin{equation}
\label{eqa:w_update}
    w_i^{(t+1)} = w_i^{(t)} \times \exp{\{
    \frac{\gamma}{n} \cdot \frac{r_i^{(t)}}{\pi_i^{(t)}}
    \}}
\end{equation}
The initial value of every $w_i^{(0)}$ is set to $1$ such that each arm is sampled with the same probability at the initial stage.

\section{Experiments}

\subsection{Datasets}

We evaluate our approach on the General Language Understanding Evaluation (GLUE) benchmark \citep{wang2018glue}, which is widely used to evaluate approaches to multi-task natural language understanding \citep{liu2019multi,clark-etal-2019-bam,Stickland:Murray:19}, as well as a number of other datasets that represent the {\em same task} as one of the GLUE datasets, but using data from a {\em different domain}. The GLUE benchmark contains nine (9) tasks, including acceptability judgements, sentiment classification, text paraphrasing, natural language inference (NLI),  question answering (QA), etc. The upper part of Table \ref{tab:datasets} presents some basic statistics about these 8 GLUE tasks (WNLI task is excluded). In addition, we select seven (7) extra datasets to evaluate how our models generalize in zero-shot and few-shot applications to new datasets. The datasets are assumed to be samples from the $\alpha$-ball within which we are minimizing the expected worst-case loss. As mentioned, each task is similar in format and objective to at least one of the original GLUE tasks. We present the data characteristics of the new datasets in the lower part of Table \ref{tab:datasets}, and list the GLUE tasks they correspond to, and whose trained classifier we will rely on in our experiments, in the last column of the Table.

\begin{table}[htbp]
\centering
% \resizebox{0.45\textwidth}{!}{%
\begin{tabular}{lc}
\toprule
\textbf{Sampling Strategy} & \textbf{Avg.} \\
\midrule
% BERT\textsubscript{BASE} & 79.6 \\
$\pi^\textsubscript{RANDOM}$ $^\dagger$&  79.1 \\
$\pi^\textsubscript{TASK SIZE}$ $^\dagger$ &  74.9  \\
$\pi^\textsubscript{EXP3.S}$ $^\dagger$ &  78.9 \\
$\pi^\textsubscript{C}$ $^\dagger$ &  80.5 \\
% PALS (pro)  &  80.6 \\
PALS (sqrt) $^\ddagger$ &  81.0 \\
PALS (anneal) $^\ddagger$ &  81.7 \\
\midrule
\textit{Our\_model} ($\phi=0$) & 83.9 \\
\textit{Our\_model} ($\phi=1$) & 84.3 \\
\textit{Our\_model} ($\phi=0.5$) & 84.2 \\
\textit{Our\_model} ($\phi=0\rightarrow 1$) & 83.8 \\
\bottomrule
\\
\end{tabular} %
% }
\caption{{\sc Comparison of Sampling Strategies} using GLUE development sets. $^\dagger$From \citet{Glover:Hokamp:19}. $^\ddagger$From \citet{Stickland:Murray:19}. The results of the four different approaches to worst-case-aware multi-task learning are not significantly different after Bonferroni correction.}
\label{tab:comparison}
\end{table}

\begin{table*}[htbp]
\centering
\resizebox{\textwidth}{!}{%
\begin{tabular}{lccccccccclc}
\toprule
    \multirow{2}{*}{\textbf{Model}}  & \textbf{CoLA} & \textbf{SST-2} & \textbf{MRPC} & \textbf{STS-B} & \textbf{QQP}  & \textbf{MNLI-m/mm} & \textbf{QNLI} & \textbf{RTE}   & \textbf{Avg.} \\
                 & 8.5k & 67k   & 3.7k & 7k    & 364k & 393k      & 108k & 2.5k & 634   &      \\
\midrule
BERT\textsubscript{BASE}$^\dagger$ & \textbf{52.1}  & \textbf{93.5} &   84.8/88.9  & 87.1/85.8 & 89.2/71.2 &  \textbf{84.6/83.4}  & 90.5 & 66.4  & 80.0 \\
MT-DNN\textsubscript{BASE}$^\ddagger$   &   44.1   &   \textbf{93.4}    &   84.6/88.8   &  85.0/84.2     &   88.4/70.5   &   84.1/83.3        &   90.8   &  \textbf{76.2}    &   79.9   \\

PALS\citep{Stickland:Murray:19} &  51.2   & 92.6 &  84.6/88.7 & 85.8/- & \textbf{89.2/71.5}  & 84.3/83.5 & 90.0 & 76.0 & 80.4 \\

$\pi^\textsubscript{C}$\citep{Glover:Hokamp:19} &  48.5 & 93.1 & 83.7/88.0  & 80.7/80.6 &   88.7/70.4 & 83.5/83.1 & 90.5 & 74.5 &  79.5 \\

\midrule

%\textit{Our\_model} ($\phi=0$) &   47.9   &   92.8    &  85.6/89.2    &  87.4/86.5     &   89.0/71.1   &   83.7/83.0       &  90.4    &   74.5    &   80.4  \\
%\textit{Our\_model} ($\phi=1$) &  46.1   &   91.8    &  84.3/88.1  &  87.5/86.6  &   88.6/70.7   &   84.2/82.8        &  \textbf{91.1}    &   74.9      &   80.0  \\
%\textit{Our\_model} (
$\phi=0.5$
&  48.0  &   92.9    &  \textbf{86.0/89.8}    &  \textbf{87.7/86.8}    &   88.8/71.2   & \textbf{84.6/83.2} &  90.9   &   75.2     &  \textbf{80.8} \\

%\textit{Our\_model} ($\phi=0\rightarrow 1$)
{\sc Anneal}-$\phi$& 48.8  &  92.7 & 85.4/89.2 &  87.3/86.4  &  88.1/70.0  & 84.5/82.8 &  \textbf{91.0}    &  75.3   &  80.6 \\

\bottomrule
\\
\end{tabular} %
}
\caption{{\sc In-Domain Results on GLUE Test Sets.} $^\dagger$Single-task training results from \citep{devlin-etal-2019-bert}.   $^\ddagger$Results of re-running the code provided by \citet{liu2019multi} without second-stage fine-tuning for the individual task. %\textbf{Bold numbers} represent best results (fine-tuned model excluded). 
We note our approach, across all settings of $\phi$, performs {\em on par} with or better than  MT-DNN\textsubscript{BASE} baselines, with generally better performance on smaller tasks. We show below that our approach also leads to much better generalization.}
\label{tab:results}
\end{table*}

\subsection{Baselines}
The baselines used for comparison in this paper all use BERT\textsubscript{BASE} as the encoder. The main differences are the sampling and training strategies used: MT-DNN \cite{liu2019multi}  uses a size-proportional sampling strategy; \citet{Glover:Hokamp:19} propose learning a sampling policy using counterfactual estimation (denoted as $\pi^\textsubscript{C}$), and compare it to uniformly random sampling $\pi^\textsubscript{RANDOM}$, size-proportional sampling ($\pi^\textsubscript{TASK SIZE}$) and automated curriculum learning ($\pi^\textsubscript{EXP3.S}$) \citep{graves2017automated}; \citet{Stickland:Murray:19} present PALS, using an annealed sampling strategy, which changes from size-proportional sampling to uniformly random sampling during the training procedure. They also compare the method to the square-root-size-proportionally sampling strategy. In Table
~\ref{tab:comparison}, we compare the different sampling strategies. Since our worst-case-aware sampling strategies are not significantly different, we hedge our bets and rely on $\phi=0.5$ and $\phi=0\rightarrow 1$ in our remaining experiments; we refer to these as $\phi=0.5$ and {\sc Anneal}-$\phi$.

\subsection{Experimental Setups}

Our implementation is based on the open-source implementation of MT-DNN.\footnote{https://github.com/namisan/mt-dnn} The original MT-DNN code adopts a size-proportional sampling strategy, but we replace this with our automated curriculum learning strategy. We also use BERT\textsubscript{BASE} as our pre-trained encoder for comparability. %, which has a moderate number of parameters and can be deployed on a single TitanX GPU. %Due to its compatibility with pytorch-transformers package\footnote{https://github.com/huggingface/transformers}, the encoder can be easily converted to other larger pre-trained modules. 
%In our MTL experiments, the weight $v_i$ of the loss for each task is simply set to $1$, as we do not specifically pay attention to one task. 
Our hyper-parameters were optimized on the GLUE development sets: We initialize the task-specific weights $w_i$ to $1$ at the beginning of each epoch. The weight $v_i$ of loss is set to $1$. The batch size is set to $8$, and the gradient accumulation step is set to $4$. The $\gamma$ value in automated curriculum learning is set to $0.001$. The number $k$ of actions in a round is set to twice the number of tasks, which is $16$. At the beginning of each round, we push one data batch per task to the queue, in case the queue of a task is empty when the trainer selects, this is excluded in the calculation of rewards. The capacity of the queue in the buffer is set to $50$. We report results with $\phi$ set to $0$, $0.5$ and $1$ respectively, as well as with increasing $\phi$ from $0$ to $1$ by $0.15$ per epoch. All other hyper-parameter values were adopted from \cite{liu2019multi} without further optimization.  %For the WNLI task, we follow \cite{liu2019multi} and \cite{devlin-etal-2019-bert} in simply predicting majority class and excluding it from training. 
Our code will be publicly available.\footnote{github.com/anonymous}

{Note, again, that our method is designed for zero-shot and few-shot transfer learning, and we are therefore mainly interested in out-of-domain performance.} In our domain transfer  experiments, we either evaluate our models directly on unseen domains or randomly select $1\%$ and $10\%$ of the target domain training set. We refer to the unsupervised domain adaptation scenario as {\em zero-shot}, and the scenarios with limited supervision as {\em few-shot}. Note our models are trained with worst-case-aware multi-task learning, but {\em not} fine-tuned for the target task. For few-shot learning, we repeat our experiments five (5) times and report average scores to reduce the variance that results from randomly subsampling the training data.  %The task-specific classifier for one task is selected according to its correlation with the GLUE task shown in Table \ref{tab:datasets}.

\subsection{In-Domain Results}

To sanity check our system, we submitted our result files to GLUE online evaluation system;\footnote{https://gluebenchmark.com/} the results are shown in Table \ref{tab:results}. The first row shows the result of fine-tuning the BERT\textsubscript{BASE} model on each single task separately, followed by the result of MT-DNN\textsubscript{BASE} model trained on all tasks. We can see that compared with single-task learning, the MTL model is comparable in terms of four tasks with large data volume (SST-2, QQP, MNLI and QNLI), and shows substantial improvement on RTE task, due to its shared encoder that contains the information from similar MNLI and QNLI tasks. However, the performance on the outlier task, such as CoLA, has been significantly undermined. This shows that MTL models with size-proportionally random sampling strategy pay more attention to dominant tasks (e.g. tasks with large amount of data) than the out-of-domain tasks, which is consistent with the findings by \cite{liu2019multi,Glover:Hokamp:19}.% After performing the second-stage fine-tuning on MT-DNN\textsubscript{BASE} for each task, we find that COLA, STS-B and MRPC tasks are significantly improved, but the improvements of MNLI and QNLI are marginal, further indicating that MTL models with size-proportional sampling bring limited benefits to outlier tasks and even inhibit their performance.

The lower part of Table \ref{tab:results} shows the results of our models using the worst-case-aware minimization strategies. It can be seen that our models are comparable with the MT-DNN model and achieve better performance on some small datasets, such as CoLA, MRPC and STS-B, which demonstrates that our sampling and training strategies have the ability to balance the training between the dominant task and the outlier task.

\begin{table*}[htbp]
\centering
\resizebox{\textwidth}{!}{ %
\begin{tabular}{llcccccccccc}
\toprule
\textbf{Setting} &\textbf{Model} & \textbf{IMDB}&  \textbf{Yelp-2} & \textbf{SICK-R} & \textbf{SICK-E} & \textbf{SNLI} & \textbf{Scitail} & \textbf{WikiQA} & \textbf{Avg.}\\
 & & 25k &  56k & 5k & 5k & 55k & 24k & 20k &\\
\midrule
% \multicolumn{11}{c}{Zero Shot (0\%)} \\ \midrule
\multirow{3}{*}{\parbox{1.3cm}{\textbf{Zero Shot (0\%)}}}&MT-DNN\textsubscript{BASE}  & 85.1 &  86.2 & 80.7/79.3 & 50.4 & 78.8 & 77.4 & 78.8/79.6 & 76.7 \\
% &MT-DNN\textsubscript{BASE} (Fine-tuned)& 85.7 &89.5 &83.8 &86.0 & 80.6/78.5 &50.4 &78.8 &76.8 & 76.7/77.8 & 78.6 \\
\cmidrule{2-10}
%&Our\_model ($\phi=0$) & 84.7 &  85.9 & 83.0/79.0 & 54.6 & 78.5 & 79.5 & 78.1/79.3 & 77.6 \\
%&Our\_model ($\phi=1$) & 85.5 &  85.7 & 82.8/79.3 & 56.8 & \textbf{80.2} & 79.2 & 79.6/80.6 &  78.4 \\
&$\phi=0.5$ & \textbf{86.2} &\textbf{88.2} & \textbf{83.0/79.4} & \textbf{57.1} & 79.7 & 80.6 & 78.5/79.3 & 78.8 \\
&{\sc Anneal}-$\phi$ & 86.1 &  87.7 & 82.1/78.7 & 56.1 &\textbf{ 79.9} & \textbf{82.3} & \textbf{81.9/82.5} & \textbf{79.3}\\
\midrule
% \multicolumn{10}{c}{Few Shot (1\%)} \\ \midrule
\multirow{3}{*}{\parbox{1.3cm}{\textbf{Few Shot (1\%)}}}
% &BERT\textsubscript{BASE}  & 82.4 &  92.3 & 55.5/58.8 & 59.6 & 79.4 & 74.2 & 54.4/55.9 & 71.5 \\

%87	91.9	82.9	79.08	82.02	85.22	87.48	79.66	80.76
&MT-DNN\textsubscript{BASE}  & \textbf{87.0} &  91.9 & 82.9/79.1 & 82.0 & \textbf{85.2} & 87.5 & 79.7/80.8 & 85.0$\pm$0.49 \\
%&Our\_model ($\phi=0$) & 86.9 &  91.9 & 85.3/79.3 & 84.9 & 84.6 & \textbf{88.1} & 78.6/79.9 & 85.4$\pm$0.24 \\

%86.8296	92.081	85.5918	79.9784	84.9814	85.02	87.3774	79.8928	80.8398
%&Our\_model ($\phi=1$) & 86.7 &  92.0 & 85.1/79.4 & 83.8 & 84.7 & 85.1 & 79.7/80.8 &  85.0\pm$0.19 \\
\cmidrule{2-10}&$\phi=0.5$ & 86.8 &  \textbf{92.1} & \textbf{85.6/80.0} & 85.0 & 85.0 & 87.4 & 79.9/80.8 & 85.6$\pm$0.32 \\
&{\sc Anneal}-$\phi$& 86.7 &  92.0 & 85.0/79.4 & \textbf{85.7} & 84.9 & \textbf{88.1} & \textbf{82.1/82.9} & \textbf{86.0$\pm$0.17} \\
\midrule
\multirow{3}{*}{\parbox{1.3cm}{\textbf{Few Shot (10\%)}}}
% &BERT\textsubscript{BASE}  & 88.2 &  94.2 & 79.2/69.9 & 75.7 & 86.6 & 89.0 & 71.4/73.1  & 83.0 \\
&MT-DNN\textsubscript{BASE}  & 88.7 &  \textbf{94.3} & 86.8/81.9 & 87.2 & {\bf 88.1} & 91.6 & 83.1/84.4 & 88.3$\pm$0.09 \\
% &Our\_model ($\phi=0$) & 88.8 &  94.1 & 87.4/82.1 & 87.2 & 87.3 & 91.3 & 82.8/83.9 & 88.1 \\
% &Our\_model ($\phi=1$) & 88.8 &  93.9 & 87.4/82.0 & 86.5 & 87.4 & 91.4 & 83.0/84.1 & 88.0 \\
\cmidrule{2-10}&$\phi=0.5$ & \textbf{89.0} &  94.2 & \textbf{87.8/82.6} & \textbf{87.4} & 87.8 & {\bf 92.4} & 83.1/84.0 & \textbf{88.5$\pm$0.10}\\

&{\sc Anneal}-$\phi$ & 88.7 &  94.2 & 87.3/82.1 & {\bf 87.4} & 87.7 & {92.0} & \textbf{84.5/85.5} & \textbf{88.5$\pm$0.14} \\
% \midrule
% \multirow{4}{*}{\parbox{1.3cm}{\textbf{Full Training (100\%)}}}&BERT\textsubscript{BASE}  & 91.6 & 95.9 & 88.8/83.3 & 85.6 & 90.9 & 94.0 & 83.5/84.8 & 89.7 \\
% & MT-DNN\textsubscript{BASE}  & 91.6 &  95.7 & 91.1/86.8 & 88.9 & 90.8 & 95.7 & 87.5/88.9 & \textbf{91.4} \\
% & Our\_model ($\phi=0$) & 91.7 &  95.7 & 91.1/86.5 & 90.1 & 90.8 & 95.1 & 87.5/88.5 & \textbf{91.5} \\
% & Our\_model ($\phi=1$) & 91.7  &  95.6 & 90.9/86.6 & 89.7 & 90.6 & 95.2 & 87.9/88.8 & \textbf{91.4} \\
% & Our\_model ($\phi=0.5$) & 91.7 &  95.7 & 91.0/86.5 & 89.4 & 90.8 & 95.1 & 87.0/88.3 & \textbf{91.3} \\
% & Our\_model ($\phi=0\rightarrow 1$) & 91.6 &  95.7 & 90.8/86.2 & 89.3 & 90.7 & 95.6 & 87.6/88.8  & \textbf{91.4}\\
\bottomrule \\
\end{tabular} %
}
\caption{{\sc Main Results (Domain Transfer).} Our models have great advantages in zero-shot and few-shot learning settings, although the gap narrows as the data volume increases. Few shot results are averages over five (5) repeated experiments to reduce the variance from random subsampling.}
\label{tab:transfer}
\end{table*}

\subsection{Domain Transfer Results}

Table \ref{tab:transfer} presents our transfer learning results after fine-tuning on 0\%~of the training data for the new task (zero shot), or 1\% or 10\%~of it (few shot). In the zero-shot setting, our models are considerably better than the MT-DNN models, across different $\phi$ values, especially for SICK-R/E and SciTail tasks. %An interesting point is that although our models are slightly inferior to the fine-tuned MT-DNN model in terms of the average performance on GLUE tasks and the sentiment classification task (SST-2), it shows better performance on additional sentiment tasks and all tasks on average. Furthermore, our model only maintains one set of parameters of encoder for all tasks, rather than one set for one task. The results on the few-shot learning settings illustrates that our models have better generalization on out-of-domain tasks than MT-DNN models.
%As for the few-shot transfer learning settings, due to a small amount of in-domain data to learn, each model has a great improvement compared to the zero-shot settings.
In the few-shot setting, we see the same trend. 
%Our performance far exceeds %Our models stand out again on these tasks, far exceeding the
%BERT\textsubscript{BASE}
%performance, and %model, even better than the one trained on 10\% data volume. When 
Compared to the MT-DNN model, our models still perform much better on small tasks (SICK-R/E, SciTail and WikiQA) and slightly better on average.
%As for the 10\% and full training settings, MTL models are still better than the single-task BERT model. However, due to the same structure of MTL models and sampling strategy for a single task, 
The gap between our models and the original MT-DNN models narrows as more training data is  available. %, as the training data grows. We can see that 
%The performance with full training is approximately the same.

\section{Analysis}

\begin{figure}[htbp]
    % \centering
    \begin{subfigure}[t]{0.5\textwidth}
        \centering
        \includegraphics[height=1.5in]{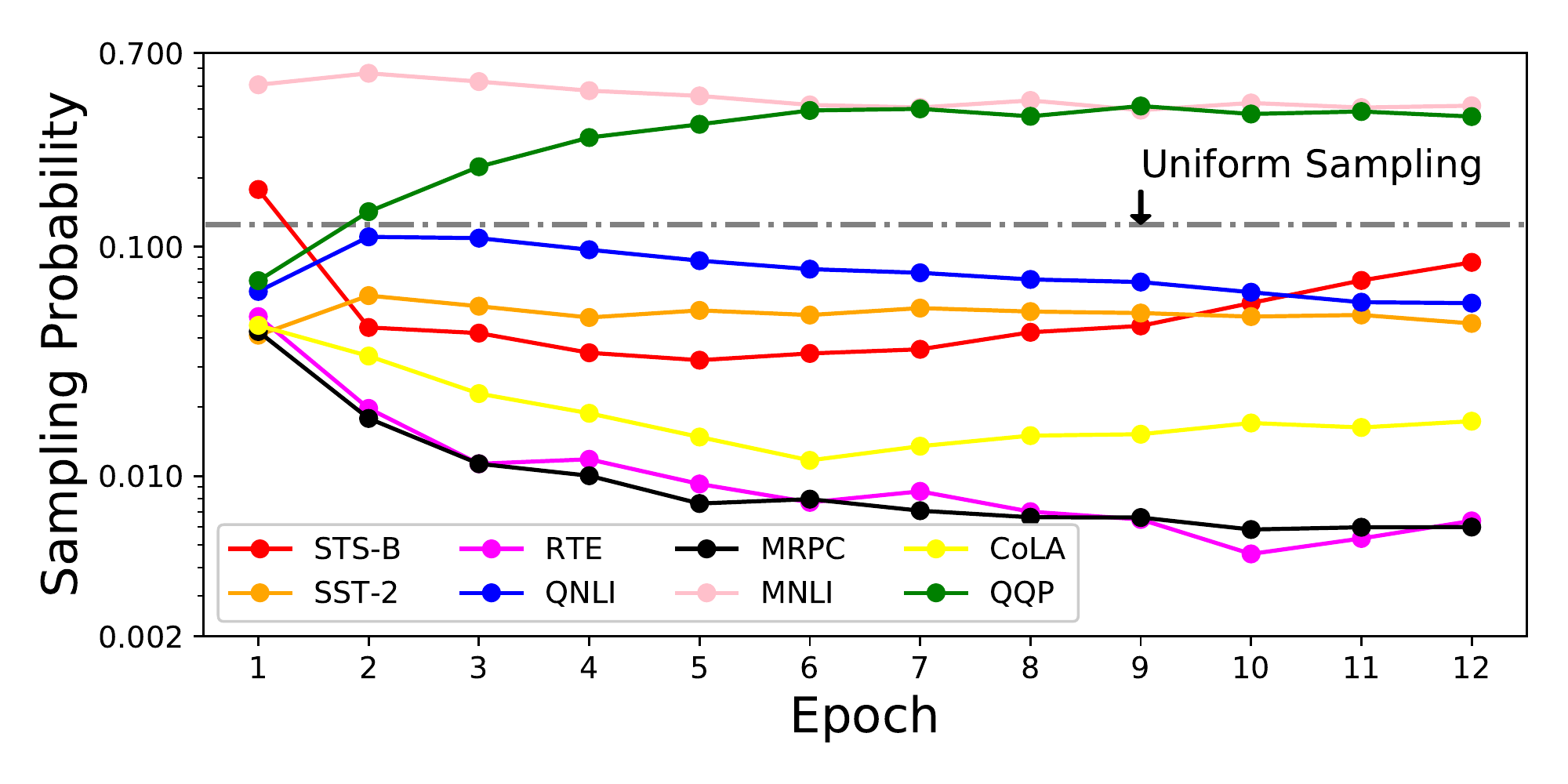}
        \caption{Sampling Probability (by Trainer), $\phi=0.5$}
        \label{fig:numerical}
    \end{subfigure}%
    \\
    \begin{subfigure}[t]{0.5\textwidth}
        \centering
        \includegraphics[height=1.5in]{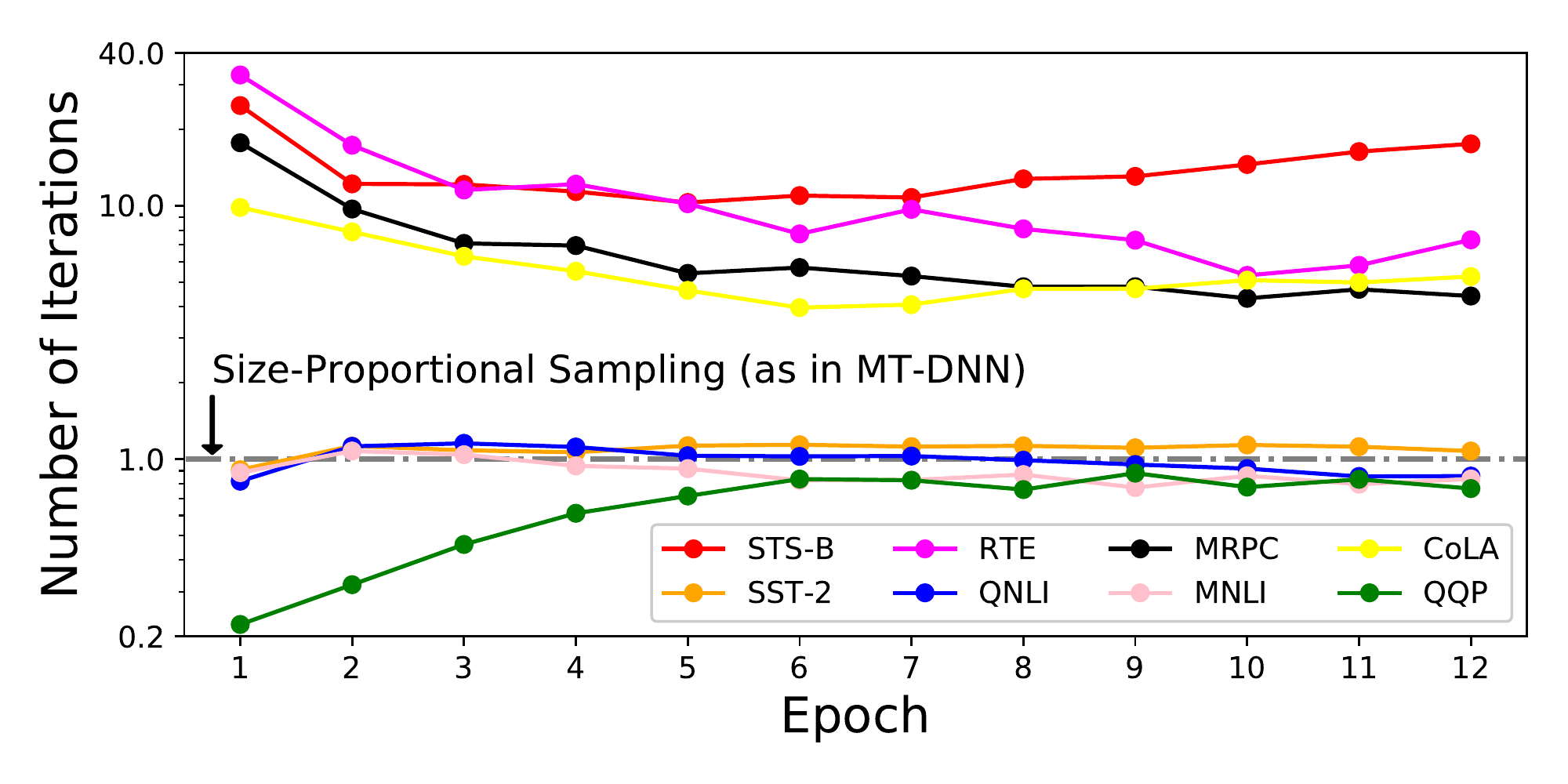}
        \caption{Number of Iterations, $\phi=0.5$}
        \label{fig:scaled}
    \end{subfigure}
    \caption{{\sc Sampling probability of tasks per epoch.} Figure (a) shows the (normalized) sampling probabilities per epoch; Figure (b) the (normalized) number of iterations per task in each epoch {\em relative to dataset size}. Two horizontal dash-dotted lines correspond to uniform (a) and size-proportional (b) sampling, respectively. %at $1.0$ in the right one indicates that every task is selected randomly according to its data volume, which is the strategy of original MT-DNN model
    Comparing (a) and (b), we see, for example, that QQP is sampled {\em more} than it would with uniform sampling, but {\em less} than it would with size-proportional sampling.}
    \label{fig:select}
\end{figure}

\begin{figure}[htbp]
    \centering
    \begin{subfigure}[t]{0.5\textwidth}
        \centering
        \includegraphics[height=1.5in]{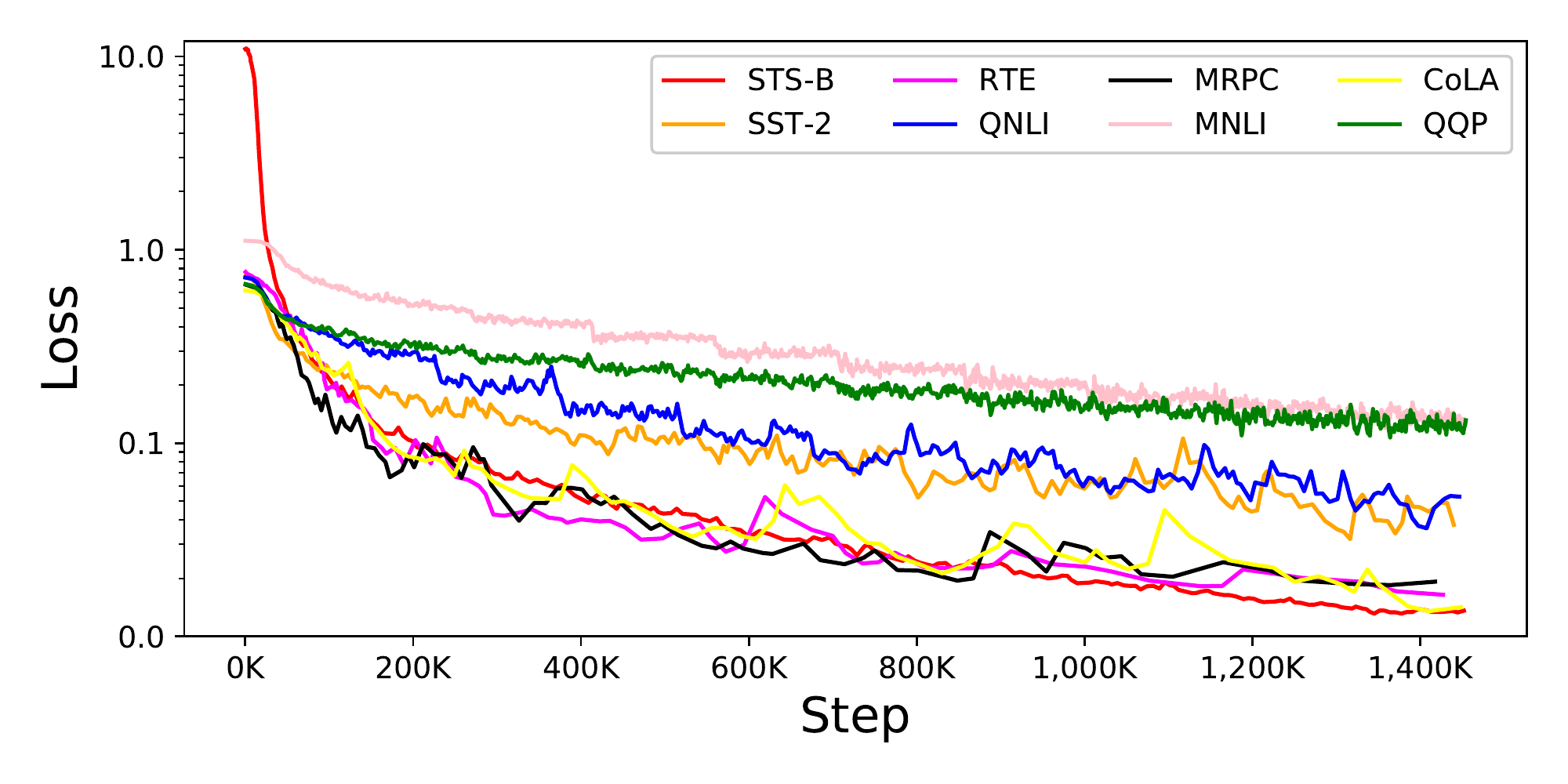}
        \caption{$\phi=0$}
        \label{fig:loss_p0}
    \end{subfigure}%
    \\
    \begin{subfigure}[t]{0.5\textwidth}
        \centering
        \includegraphics[height=1.5in]{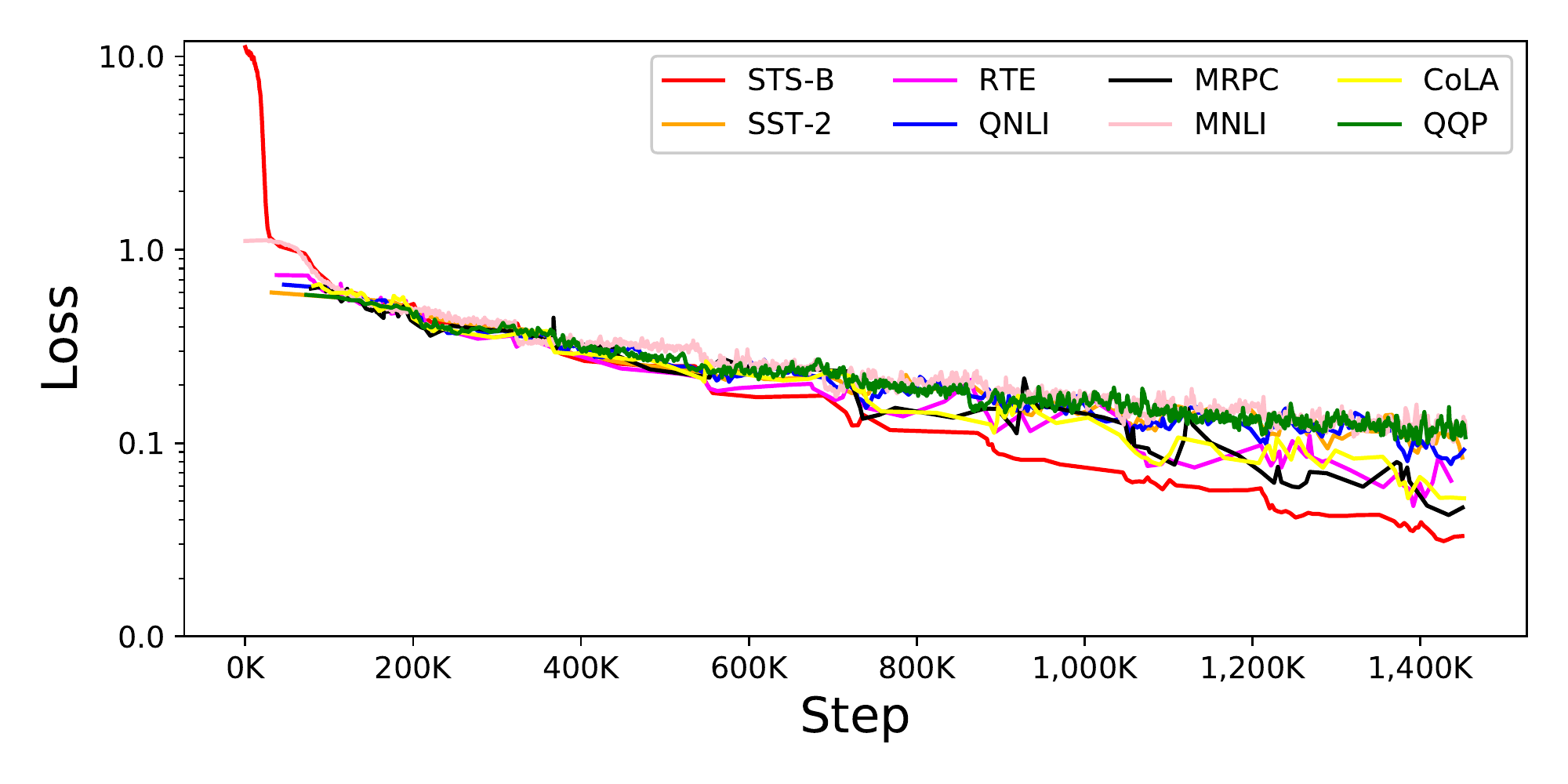}
        \caption{$\phi=1$}
        \label{fig:loss_p1}
    \end{subfigure}
    \caption{{\sc Loss curves.} Strict worst-case minimization (b) synchronizes loss curves.  %The moving average is performed on loss values for better presentation of the trend. Curves on the left are sparse due to possibility that every task can be selected, whereas on the right are compact since the strategy only cares about the worst case. At most steps, loss-proportional sampling ($\phi=0$) models have higher maximum but lower average of losses than worst-case sampling ($\phi=1$) models.
    }
    \label{fig:loss}
\end{figure}

Figure \ref{fig:select} shows statistics of each task selected by the trainer during the training process. In Figure \ref{fig:numerical}, the sampling probability of each task is initially proportional to the size of the training data  (except for the STS-B, which uses numerically larger mean squared error loss; see Figure \ref{fig:loss}). When we inspect the number of times a task was selected (iterations) across epochs, curves cluster in two groups of four (4) tasks; see Figure
~\ref{fig:select}b. Tasks in the lower cluster have relatively large amounts of data, and the curves gradually converge to around 1. On the contrary, tasks in the upper cluster have small amounts of data, and their datasets are queried several times per epoch. Note that our approach does not explicitly take dataset size into account.

Figure \ref{fig:loss} shows the trend of training loss of each task when $\phi$ is set to two extremes 0 and 1. If $\phi$ equals 0, the task is sampled loss-proportionally. The loss for some tasks drops quickly, but they can still be selected by the trainer, therefore curves are apart from each other. On the contrary, the trainer selects the worst-case task when $\phi$ equals 1. Once the loss of a task is not the maximum, it will not be selected by the trainer, as a result, the curves are close to each other. %Although the maximum loss when $\phi = 1$ is generally smaller than when $\phi = 1$ at the same step, the former model has the better overall performance on GLUE benchmark. This may explain why our models with $\phi=0.5$ or changing $\phi$ have better performance on GLUE benchmark: our hybrid strategies make the upper bound (maximum loss) lower, and other losses at the same time are apart from the upper bound.

\section{Related Work}

\paragraph{Sampling strategies for multi-task learning} While most previous work in multi-task learning relies on uniform or proportional sampling, more sophisticated sampling techniques have been proposed.  \citet{Stickland:Murray:19} presents square root sampling, which balances uniform and proportional sampling, as well as annealed sampling, which gradually moves from one to the other across training epochs. \citet{xu-etal-2019-multi} use language models to assign weights to data before sampling. None of this work evaluates the robustness of sampling strategies in domain transfer scenarios.

\paragraph{Multi-task curricula} \citet{Guo:ea:19} use Bayesian optimization to learn a fixed curriculum for multi-task learning problems derived from the GLUE benchmark. \citet{Glover:Hokamp:19} evaluate fixed curricula, as well as the automated curriculum learning approach in \citet{graves2017automated}, on the GLUE benchmark and show how this approach can be improved by counterfactual policy estimation. \citet{Zaremoodi:Haffari:20}, in contrast, formulate the curricular as a Markov decision process and adopt oracle policy to adaptively and dynamically select different tasks.

\paragraph{Robust curricula} We are not the first to use automated curriculum learning to minimize a worst-case loss. \citet{Cai:ea:18} learn curricula of adversarial examples generated by attacks. \citet{sharma2018learning} also present three worst-case-aware algorithms for learning multi-task curricula, which they evaluate on stochastic computer games, quite different from the GLUE benchmark: Two of them use multi-armed bandits to update the sampling distribution by the end of each episode. The reward used to train the multi-armed bandit is the normalized task performance of the three (currently) worst tasks. Apart from the fact that the learning problem considered here is very different from theirs, our approach also differs from theirs in that we use a more general loss function instead of relying on the top-3 heuristic. 

\section{Conclusion}

In this work, we introduce a worst-case-aware approach to automated curriculum learning for zero-shot and few-shot applications of multi-task learning architectures. Our models achieves competitive or slightly better average performance on the GLUE benchmark and, more importantly, improves %the performance on outlier tasks. Furthermore, we showed that our worst-case-aware strategies have better 
generalization to out-of-domain tasks in zero-shot and few-shot  %learning-to-learn
settings. Our approach also generally leads to better performance on small tasks. We analyze the learning dynamics of automated curriculum learning in this context and show how it learns a very different sampling strategy from commonly used baseline heuristics. 

\bibliography{refs}

\begin{thebibliography}{33}
\providecommand{\natexlab}[1]{#1}
\providecommand{\url}[1]{\texttt{#1}}
\providecommand{\urlprefix}{URL }
\expandafter\ifx\csname urlstyle\endcsname\relax
  \providecommand{\doi}[1]{doi:\discretionary{}{}{}#1}\else
  \providecommand{\doi}{doi:\discretionary{}{}{}\begingroup
  \urlstyle{rm}\Url}\fi

\bibitem[{Baxter(2000)}]{Baxter:00}
Baxter, J. 2000.
\newblock A model of inductive bias learning.
\newblock \emph{Journal of Artificial Intelligence Research} 12.

\bibitem[{Bowman et~al.(2015)Bowman, Angeli, Potts, and
  Manning}]{bowman-etal-2015-large}
Bowman, S.~R.; Angeli, G.; Potts, C.; and Manning, C.~D. 2015.
\newblock A large annotated corpus for learning natural language inference.
\newblock In \emph{Proceedings of the 2015 Conference on Empirical Methods in
  Natural Language Processing}, 632--642. Lisbon, Portugal: Association for
  Computational Linguistics.
\newblock \doi{10.18653/v1/D15-1075}.
\newblock \urlprefix\url{https://www.aclweb.org/anthology/D15-1075}.

\bibitem[{Bronskill et~al.(2020)Bronskill, Gordon, Requeima, Nowozin, and
  Turner}]{bronskill2020tasknorm}
Bronskill, J.; Gordon, J.; Requeima, J.; Nowozin, S.; and Turner, R.~E. 2020.
\newblock TaskNorm: Rethinking Batch Normalization for Meta-Learning.
\newblock \emph{arXiv preprint arXiv:2003.03284} .

\bibitem[{Bubeck, Munos, and Stoltz(2008)}]{bubeck2008pure}
Bubeck, S.; Munos, R.; and Stoltz, G. 2008.
\newblock Pure Exploration for Multi-Armed Bandit Problems.
\newblock \emph{arXiv preprint arXiv:2004.07162} .

\bibitem[{Cai, Liu, and Song(2018)}]{Cai:ea:18}
Cai, Q.-Z.; Liu, C.; and Song, D. 2018.
\newblock Curriculum adversarial training.
\newblock In \emph{Proceedings of the 27th International Joint Conference on
  Artificial Intelligence}, 3740--3747.

\bibitem[{Caruana(1997)}]{Caruana:97}
Caruana, R. 1997.
\newblock Multitask Learning.
\newblock \emph{Mach. Learn.} 28(1): 41–75.
\newblock ISSN 0885-6125.
\newblock \doi{10.1023/A:1007379606734}.
\newblock \urlprefix\url{https://doi.org/10.1023/A:1007379606734}.

\bibitem[{Chen et~al.(2018)Chen, Badrinarayanan, Lee, and
  Rabinovich}]{chen2018gradnorm}
Chen, Z.; Badrinarayanan, V.; Lee, C.-Y.; and Rabinovich, A. 2018.
\newblock GradNorm: Gradient Normalization for Adaptive Loss Balancing in Deep
  Multitask Networks.
\newblock In \emph{International Conference on Machine Learning}, 794--803.

\bibitem[{Clark et~al.(2019)Clark, Luong, Khandelwal, Manning, and
  Le}]{clark-etal-2019-bam}
Clark, K.; Luong, M.-T.; Khandelwal, U.; Manning, C.~D.; and Le, Q.~V. 2019.
\newblock {BAM}! Born-Again Multi-Task Networks for Natural Language
  Understanding.
\newblock In \emph{Proceedings of the 57th Annual Meeting of the Association
  for Computational Linguistics}, 5931--5937. Florence, Italy: Association for
  Computational Linguistics.
\newblock \doi{10.18653/v1/P19-1595}.
\newblock \urlprefix\url{https://www.aclweb.org/anthology/P19-1595}.

\bibitem[{Devlin et~al.(2019)Devlin, Chang, Lee, and
  Toutanova}]{devlin-etal-2019-bert}
Devlin, J.; Chang, M.-W.; Lee, K.; and Toutanova, K. 2019.
\newblock {BERT}: Pre-training of Deep Bidirectional Transformers for Language
  Understanding.
\newblock In \emph{Proceedings of the 2019 Conference of the North {A}merican
  Chapter of the Association for Computational Linguistics: Human Language
  Technologies, Volume 1 (Long and Short Papers)}, 4171--4186. Minneapolis,
  Minnesota: Association for Computational Linguistics.
\newblock \doi{10.18653/v1/N19-1423}.
\newblock \urlprefix\url{https://www.aclweb.org/anthology/N19-1423}.

\bibitem[{Duchi, Hashimoto, and Namkoong(2020)}]{Duchi:ea:20}
Duchi, J.~C.; Hashimoto, T.; and Namkoong, H. 2020.
\newblock Distributionally Robust Losses Against Mixture Covariate Shifts.
\newblock In \emph{Unpublished manuscript}.

\bibitem[{Glover and Hokamp(2019)}]{Glover:Hokamp:19}
Glover, J.; and Hokamp, C. 2019.
\newblock Task Selection Policies in Multitask Learning.
\newblock \emph{arXiv preprint arXiv:1907.06214} .

\bibitem[{Graves et~al.(2017)Graves, Bellemare, Menick, Munos, and
  Kavukcuoglu}]{graves2017automated}
Graves, A.; Bellemare, M.~G.; Menick, J.; Munos, R.; and Kavukcuoglu, K. 2017.
\newblock Automated curriculum learning for neural networks.
\newblock In \emph{Proceedings of the 34th International Conference on Machine
  Learning-Volume 70}, 1311--1320. JMLR. org.

\bibitem[{Guo, Pasunuru, and Bansal(2019)}]{Guo:ea:19}
Guo, H.; Pasunuru, R.; and Bansal, M. 2019.
\newblock AutoSeM: Automatic Task Selection and Mixing in Multi-Task Learning.
\newblock In \emph{Proceedings of the 2019 Conference of the North American
  Chapter of the Association for Computational Linguistics: Human Language
  Technologies, Volume 1 (Long and Short Papers)}, 3520--3531.

\bibitem[{Hashimoto et~al.(2018)Hashimoto, Srivastava, Namkoong, and
  Liang}]{pmlr-v80-hashimoto18a}
Hashimoto, T.; Srivastava, M.; Namkoong, H.; and Liang, P. 2018.
\newblock Fairness Without Demographics in Repeated Loss Minimization.
\newblock In \emph{International Conference on Machine Learning}, 1929--1938.

\bibitem[{Herbster and Warmuth(1998)}]{journals/ml/HerbsterW98}
Herbster, M.; and Warmuth, M.~K. 1998.
\newblock Tracking the Best Expert.
\newblock \emph{Mach. Learn.} 32(2): 151--178.
\newblock
  \urlprefix\url{http://dblp.uni-trier.de/db/journals/ml/ml32.html#HerbsterW98}.

\bibitem[{Hernandez-Lobato, Hernandez-Lobato, and
  Ghahramani(2015)}]{pmlr-v37-hernandez-lobatoa15}
Hernandez-Lobato, D.; Hernandez-Lobato, J.~M.; and Ghahramani, Z. 2015.
\newblock A Probabilistic Model for Dirty Multi-task Feature Selection.
\newblock In Bach, F.; and Blei, D., eds., \emph{Proceedings of the 32nd
  International Conference on Machine Learning}, volume~37 of \emph{Proceedings
  of Machine Learning Research}, 1073--1082. Lille, France: PMLR.
\newblock
  \urlprefix\url{http://proceedings.mlr.press/v37/hernandez-lobatoa15.html}.

\bibitem[{Khot, Sabharwal, and Clark(2018)}]{khot2018scitail}
Khot, T.; Sabharwal, A.; and Clark, P. 2018.
\newblock Scitail: A textual entailment dataset from science question
  answering.
\newblock In \emph{Thirty-Second AAAI Conference on Artificial Intelligence}.

\bibitem[{Kuhn et~al.(2019)Kuhn, Esfahani, Nguyen, and
  Shafieezadeh{-}Abadeh}]{DBLP:journals/corr/abs-1908-08729}
Kuhn, D.; Esfahani, P.~M.; Nguyen, V.~A.; and Shafieezadeh{-}Abadeh, S. 2019.
\newblock Wasserstein Distributionally Robust Optimization: Theory and
  Applications in Machine Learning.
\newblock \emph{CoRR} abs/1908.08729.
\newblock \urlprefix\url{http://arxiv.org/abs/1908.08729}.

\bibitem[{Liu et~al.(2019)Liu, He, Chen, and Gao}]{liu2019multi}
Liu, X.; He, P.; Chen, W.; and Gao, J. 2019.
\newblock Multi-Task Deep Neural Networks for Natural Language Understanding.
\newblock In \emph{Proceedings of the 57th Annual Meeting of the Association
  for Computational Linguistics}, 4487--4496.

\bibitem[{Maas et~al.(2011)Maas, Daly, Pham, Huang, Ng, and
  Potts}]{maas2011learning}
Maas, A.~L.; Daly, R.~E.; Pham, P.~T.; Huang, D.; Ng, A.~Y.; and Potts, C.
  2011.
\newblock Learning word vectors for sentiment analysis.
\newblock In \emph{Proceedings of the 49th annual meeting of the association
  for computational linguistics: Human language technologies-volume 1},
  142--150. Association for Computational Linguistics.

\bibitem[{Marelli et~al.(2014)Marelli, Menini, Baroni, Bentivogli, Bernardi,
  Zamparelli et~al.}]{marelli2014sick}
Marelli, M.; Menini, S.; Baroni, M.; Bentivogli, L.; Bernardi, R.; Zamparelli,
  R.; et~al. 2014.
\newblock A SICK cure for the evaluation of compositional distributional
  semantic models.
\newblock In \emph{LREC}, 216--223.

\bibitem[{Mehta, Lee, and Gray(2012)}]{mehta2012minimax}
Mehta, N.; Lee, D.; and Gray, A.~G. 2012.
\newblock Minimax multi-task learning and a generalized loss-compositional
  paradigm for MTL.
\newblock In \emph{Advances in Neural Information Processing Systems},
  2150--2158.

\bibitem[{Oren et~al.(2019)Oren, Sagawa, Hashimoto, and
  Liang}]{oren-etal-2019-distributionally}
Oren, Y.; Sagawa, S.; Hashimoto, T.; and Liang, P. 2019.
\newblock Distributionally Robust Language Modeling.
\newblock In \emph{Proceedings of the 2019 Conference on Empirical Methods in
  Natural Language Processing and the 9th International Joint Conference on
  Natural Language Processing (EMNLP-IJCNLP)}, 4227--4237. Hong Kong, China:
  Association for Computational Linguistics.
\newblock \doi{10.18653/v1/D19-1432}.
\newblock \urlprefix\url{https://www.aclweb.org/anthology/D19-1432}.

\bibitem[{Sharma et~al.(2018)Sharma, Jha, Hegde, and
  Ravindran}]{sharma2018learning}
Sharma, S.; Jha, A.~K.; Hegde, P.~S.; and Ravindran, B. 2018.
\newblock Learning to Multi-Task by Active Sampling.
\newblock In \emph{International Conference on Learning Representations}.
\newblock \urlprefix\url{https://openreview.net/forum?id=B1nZ1weCZ}.

\bibitem[{Stickland and Murray(2019)}]{Stickland:Murray:19}
Stickland, A.~C.; and Murray, I. 2019.
\newblock BERT and PALs: Projected Attention Layers for Efficient Adaptation in
  Multi-Task Learning.
\newblock In \emph{ICML}.

\bibitem[{Wang et~al.(2018)Wang, Singh, Michael, Hill, Levy, and
  Bowman}]{wang2018glue}
Wang, A.; Singh, A.; Michael, J.; Hill, F.; Levy, O.; and Bowman, S. 2018.
\newblock {GLUE}: A Multi-Task Benchmark and Analysis Platform for Natural
  Language Understanding.
\newblock In \emph{Proceedings of the 2018 {EMNLP} Workshop {B}lackbox{NLP}:
  Analyzing and Interpreting Neural Networks for {NLP}}, 353--355. Brussels,
  Belgium: Association for Computational Linguistics.
\newblock \doi{10.18653/v1/W18-5446}.
\newblock \urlprefix\url{https://www.aclweb.org/anthology/W18-5446}.

\bibitem[{Wolpert(1996)}]{DBLP:journals/neco/Wolpert96}
Wolpert, D.~H. 1996.
\newblock The Lack of {A} Priori Distinctions Between Learning Algorithms.
\newblock \emph{Neural Computation} 8(7): 1341--1390.
\newblock \doi{10.1162/neco.1996.8.7.1341}.
\newblock \urlprefix\url{https://doi.org/10.1162/neco.1996.8.7.1341}.

\bibitem[{Xu et~al.(2019)Xu, Liu, Shen, Liu, and Gao}]{xu-etal-2019-multi}
Xu, Y.; Liu, X.; Shen, Y.; Liu, J.; and Gao, J. 2019.
\newblock Multi-task Learning with Sample Re-weighting for Machine Reading
  Comprehension.
\newblock In \emph{Proceedings of the 2019 Conference of the North {A}merican
  Chapter of the Association for Computational Linguistics: Human Language
  Technologies, Volume 1 (Long and Short Papers)}, 2644--2655. Minneapolis,
  Minnesota: Association for Computational Linguistics.
\newblock \doi{10.18653/v1/N19-1271}.
\newblock \urlprefix\url{https://www.aclweb.org/anthology/N19-1271}.

\bibitem[{Yang, Yih, and Meek(2015)}]{yang-etal-2015-wikiqa}
Yang, Y.; Yih, W.-t.; and Meek, C. 2015.
\newblock {W}iki{QA}: A Challenge Dataset for Open-Domain Question Answering.
\newblock In \emph{Proceedings of the 2015 Conference on Empirical Methods in
  Natural Language Processing}, 2013--2018. Lisbon, Portugal: Association for
  Computational Linguistics.
\newblock \doi{10.18653/v1/D15-1237}.
\newblock \urlprefix\url{https://www.aclweb.org/anthology/D15-1237}.

\bibitem[{Yue, Kuhn, and Wiesemann(2020)}]{yue2020linear}
Yue, M.-C.; Kuhn, D.; and Wiesemann, W. 2020.
\newblock On Linear Optimization over Wasserstein Balls.
\newblock \emph{arXiv preprint arXiv:2004.07162} .

\bibitem[{Zaremoodi and Haffari(2020)}]{Zaremoodi:Haffari:20}
Zaremoodi, P.; and Haffari, G. 2020.
\newblock Learning to Multi-Task Learn for Better Neural Machine Translation.
\newblock \emph{arXiv preprint arXiv:2001.03294} .

\bibitem[{Zhang, Yeung, and Xu(2010)}]{NIPS2010_4150}
Zhang, Y.; Yeung, D.-Y.; and Xu, Q. 2010.
\newblock Probabilistic Multi-Task Feature Selection.
\newblock In Lafferty, J.~D.; Williams, C. K.~I.; Shawe-Taylor, J.; Zemel,
  R.~S.; and Culotta, A., eds., \emph{Advances in Neural Information Processing
  Systems 23}, 2559--2567. Curran Associates, Inc.
\newblock
  \urlprefix\url{http://papers.nips.cc/paper/4150-probabilistic-multi-task-feature-selection.pdf}.

\bibitem[{Zhou et~al.(2020)Zhou, Yang, Hospedales, and
  Xiang}]{DBLP:journals/corr/abs-2003-06054}
Zhou, K.; Yang, Y.; Hospedales, T.~M.; and Xiang, T. 2020.
\newblock Deep Domain-Adversarial Image Generation for Domain Generalisation.
\newblock \emph{CoRR} abs/2003.06054.
\newblock \urlprefix\url{https://arxiv.org/abs/2003.06054}.

\end{thebibliography}

\end{document}